\documentclass[10pt,twocolumn,letterpaper]{article}

\usepackage[pagenumbers]{wacv} 

\usepackage{graphicx}
\usepackage{amsmath}
\usepackage{amssymb}
\usepackage{booktabs}
\usepackage{algorithm}
\usepackage{algpseudocode}
\usepackage{relsize}
\usepackage{hhline}
\usepackage{caption}
\usepackage{pifont}
\usepackage{makecell}
\usepackage{xcolor}

\usepackage[pagebackref,breaklinks,colorlinks]{hyperref}

\usepackage[capitalize]{cleveref}
\crefname{section}{Sec.}{Secs.}
\Crefname{section}{Section}{Sections}
\Crefname{table}{Table}{Tables}
\crefname{table}{Tab.}{Tabs.}

\providecommand{\bt}[1]{\textcolor{blue}{#1}}
\providecommand{\gt}[1]{\textcolor{green}{#1}}
\providecommand{\rt}[1]{\textcolor{red}{#1}}

\newcolumntype{?}{!{\vrule width 1.2pt}}

\def\wacvPaperID{2218} 
\def\confName{WACV}
\def\confYear{2025}

\begin{document}

\title{Mamba-ST: State Space Model for Efficient Style Transfer}
\author{Filippo Botti
\qquad
Alex Ergasti
\qquad
Leonardo Rossi
\qquad
Tomaso Fontanini\\
\qquad
Claudio Ferrari
\qquad
Massimo Bertozzi
\qquad
Andrea Prati\\
University of Parma, Department of Engineering and Architecture\\
Parma, Italy\\
{\tt\small \{filippo.botti,  alex.ergasti, leonardo.rossi, claudio.ferrari2,}\\\tt\small{ massimo.bertozzi, andrea.prati\}@unipr.it} 
}

\maketitle

\begin{abstract}
The goal of style transfer is, given a content image and a style source, generating a new image preserving the content but with the artistic representation of the style source. 
Most of the state-of-the-art architectures use transformers or diffusion-based models to perform this task, despite the heavy computational burden that they require. In particular, transformers use self- and cross-attention layers which have large memory footprint, while diffusion models require high inference time.
To overcome the above, this paper explores a novel design of Mamba, an emergent State-Space Model (SSM), called Mamba-ST, to perform style transfer.
To do so, we adapt Mamba linear equation to simulate the behavior of cross-attention layers, which are able to combine two separate embeddings into a single output, but drastically reducing memory usage and time complexity. We modified the Mamba's inner equations so to accept inputs from, and combine, two separate data streams.
To the best of our knowledge, this is the first attempt to adapt the equations of SSMs to a vision task like style transfer without requiring any other module like cross-attention or custom normalization layers. 
An extensive set of experiments demonstrates the superiority and efficiency of our method in performing style transfer compared to transformers and diffusion models. Results show improved quality in terms of both ArtFID and FID metrics.
Code is available at \url{https://github.com/FilippoBotti/MambaST}. 
\end{abstract}


\section{Introduction}
\begin{figure}
    \centering
    \includegraphics[width=\linewidth]{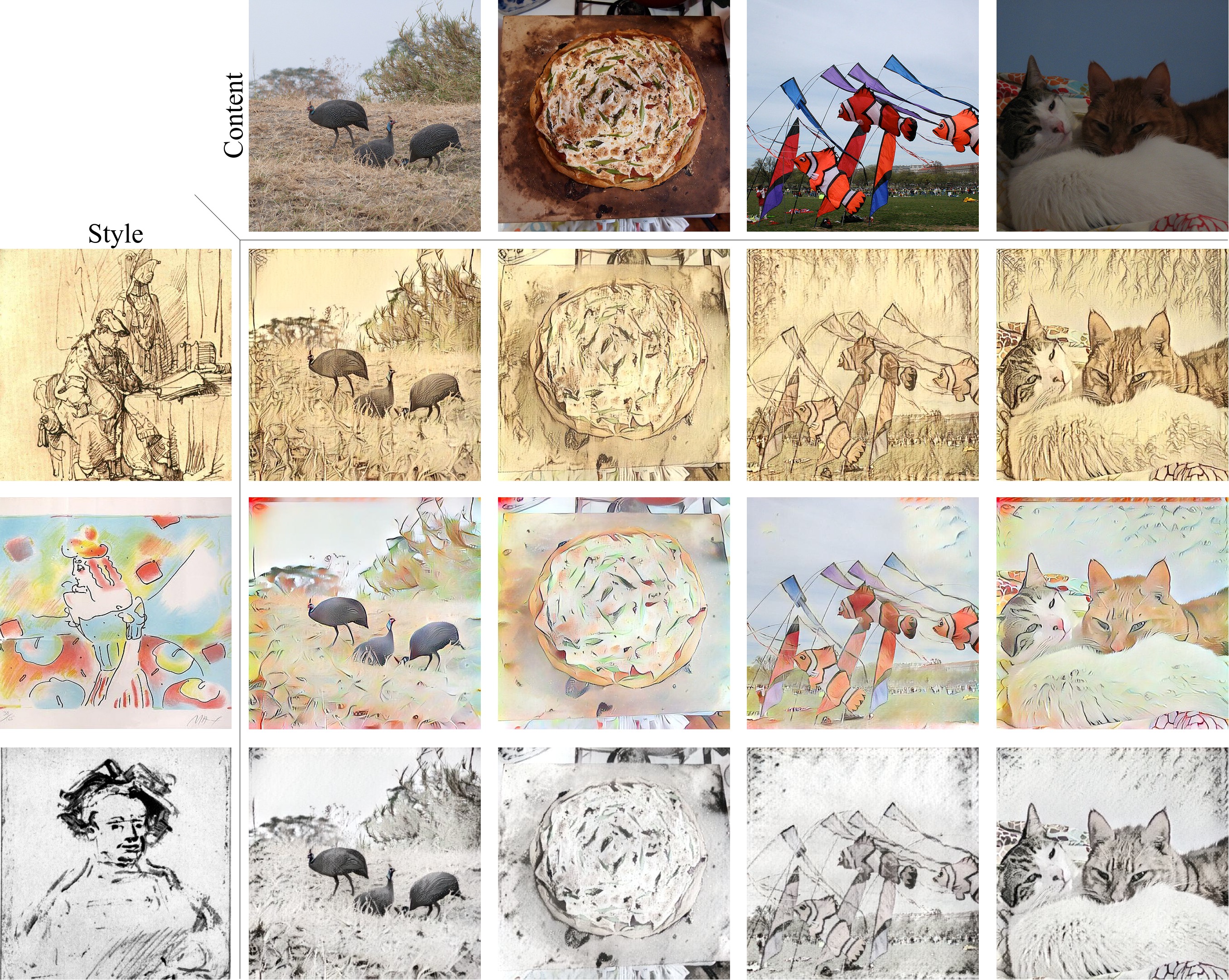}
    \caption{Examples of generated images from our Mamba model given a style and a content image}
    \label{fig:gen_images}
\end{figure}
\label{sec:intro}

Style Transfer is a deep learning technique aiming to generate a new image which has the content (\textit{e.g} objects, layout) of a given image (\textit{i.e.} the content image) and the style (\textit{e.g.} color or texture structure) of another image (\textit{i.e.} the style image). Style Transfer has been largely studied \cite{deng2022stytr2,chung2024style,kwon2022clipstyler,kim2022diffusionclip,huang2017arbitrary,li2017universal,wang2020diversified, lu2019closed} and there exist several models which can perform it with good results in terms of quality and consistency between content and style. 
Recently, style transfer was also extended to text-based models by substituting the style image with a textual description \cite{kwon2022clipstyler, kim2022diffusionclip}. In this work, we will not tackle or compare with this category of models as they largely differ from standard style transfer architectures.
General approaches follow the encoder-decoder pipeline and perform style transfer in an intermediate layer between these two sections \cite{huang2017arbitrary,li2017universal,wang2020diversified,lu2019closed}. However, these models struggle to find a relation between content and style, leading to poor quality results in terms of image details. For this reason, recent architectures (\textit{e.g.} \cite{deng2022stytr2}) leverage the capability of transformers by taking advantage of cross-attention mechanism to improve the quality of the results. However, transformer-based architectures have a large memory requirement.
On the other hand, several approaches \cite{chung2024style, wang2023stylediffusion,zhang2023inversion} tried to leverage diffusion models as backbones for style transfer, but did not reach the same quality as transformer-based architecture or, as stated in \cite{chung2024style}, required high inference time compared to the former.

On the contrary, State Space Models (SSM) \cite{gu2022efficiently,gu2021combining} have recently shown results comparable to transformers for long sequence modeling and in vision task \cite{zhu2024vision,guo2024mambair, hu2024zigma}. In particular, among these, Mamba \cite{mamba} proved to be a competitive alternative to transformers, 
but with a much lower memory requirement. Additionally, Mamba complexity scales linearly with the sequence length, rather than quadratically, resulting in a fast inference especially if compared with diffusion models. Mamba was initially designed to work with 1D sequences (like words in a sentence), but was recently adapted to work with 2D vision data thanks to VMamba \cite{liu2024vmambavisualstatespace}. Recently, in a text style transfer architecture described by Wang \etal \cite{wang2024stylemamba}, Mamba building blocks demonstrated performance comparable to transformer-based architectures. However, additional Adaptive Layer Normalization (AdaLN) was necessary to fuse the style and content.

In this work, we propose a way to adapt the inner equations of Mamba to perform style transfer (Fig.~\ref{fig:gen_images}) with a novel architecture called Mamba-ST (Mamba Style Transfer). 
The main part of Mamba-ST is a novel block, called Mamba-ST Decoder (MSTD), which is able to fuse the style information extracted from an image with the content of a different image. More in detail, both content and style are modelled as a sequence of patch embeddings and are fed to our proposed MSTD. 
In order to perform the fusion, we modified the Mamba internal matrices to mimic the functionality of a cross-attention layer, yet maintaining the core properties of SSMs. By doing so, our solution enables the interaction between style and content image without the need of additional modules, like Adaptive Layer Normalization (AdaLN). Our contributions can be summarized as:
\begin{itemize}
    \item We designed a cross-attention-like method inside SSMs. This is done by adapting the mathematical formulation of internal matrices so that additional layers like AdaLN are not required, whilst maintaining the same properties of the basic Mamba block.
    \item A novel vision-based Mamba architecture, called Mamba-ST, which is able to perform style transfer with comparable results with respect to transformer- and diffusion-based architectures.
    \item The proposed approach allows a better memory usage w.r.t. transformers and a much faster inference time compared with diffusion models.
\end{itemize}



\section{Related Work}

\paragraph{Style Transfer.}
The problem of style transfer has been widely studied in the literature \cite{huang2017arbitrary, kwon2022diffusion,deng2020arbitrary,liu2021adaattn,an2021artflow,chandran2021adaptive,hong2023aespa}. The first attempt to transfer style was the one proposed by Gatys \etal \cite{gatys2016image}, which shows that is possible to merge content and style features extracted by a  CNN by solving an optimization problem. Later, the introduction of AdaIN \cite{huang2017arbitrary} allowed to perform arbitrary style transfer adjusting the mean and the variance of the content image and align them with the ones of the style image. Thanks to its efficiency, AdaIN became a very popular architecture. Later, the advent of transformers \cite{liu2021adaattn, park2019arbitrary} showed how self-attention mechanism can improve the quality of the results by finding stronger relations between the style and the content. Subsequently, Deng \etal \cite{deng2022stytr2} introduced a fully-transformer-based architecture, StyTr$^2$, which, combined with a new content-aware positional encoding scheme, outperforms state-of-the-art methods for style transfer. Despite the capability reached in terms of quality, these methods heavily depend on transformers, so they scale quadratically with the image size, which limits their use only on small images.

Recently, diffusion-based style transfer methods \cite{zhang2023inversion, wang2023stylediffusion, chung2024style} showed how to leverage the generative capability of diffusion models in order to perform style transfer.
InST \cite{zhang2023inversion} captured the information of the style with a text-based inversion method and then transfer it.
StyleDiffusion \cite{wang2023stylediffusion} introduced a CLIP-based style disentanglement loss to disentangle style and content in the CLIP image space. Despite the quality of the images produced with these architectures, the content and the style are not perfectly merged together and the results are not yet on par with the ones generated by transformer-based models.
Finally, StyleID \cite{chung2024style} proposed a new style transfer method which exploits the knowledge of Stable Diffusion 1.4 \cite{rombach2022high} without requiring supervision or optimisation. StyleID simply substitutes key and value of content with those of styles inside self-attention layers, but, despite the improved results w.r.t. previous methods, it still requires high inference time compared to the one of transformer-based architectures. In order to solve the mentioned problems regarding memory usage and speed, while maintaining quality and coherency during style transfer, we propose a new full Mamba-based architecture.

\paragraph{State Space Models.}
Despite the well known superiority of transformer-based architecture in vision tasks \cite{DBLP:journals/corr/abs-2010-11929,carion2020end,sun2019videobert, yang2020learning,ye2019cross}, one of the most crucial problems of these architecture remains their quadratic complexity and high memory requirement. For this reason, recently, several works tried to overcome this issue \cite{dao2022flashattention,choromanski2020rethinking,ding2023longnet,wang2020linformer,dao2023flashattention}. In particular, State Space Models (SSMs) \cite{gu2022efficiently, gu2021combining} were inspired by control systems theory and have been recently introduced in deep learning field in order to take the advantage of their linear complexity \cite{hu2024zigma, wang2024mambabyte, zhu2024vision, liu2024vmambavisualstatespace}. Structured State Space Models \cite{gu2022efficiently} proposed a new parametrization for SSM in order to get the advantage of parallelization during training and achieve high speed during inference. 
Mamba \cite{mamba} was recently introduced as an improved SSM. Its main contribution consists in making the SSM parameters input-dependent. Since its superiority compared to other SSMs in terms of memory usage, time complexity and quality of the results, Mamba has been widely applied in deep learning, from NLP \cite{wang2024mambabyte}, to vision field \cite{zhu2024vision, liu2024vmambavisualstatespace} like super-resolution \cite{shi2024vmambair, guo2024mambair} or even diffusion models \cite{hu2024zigma}. Recently, a first attempt of using Mamba for text-driven style transfer was presented \cite{wang2024stylemamba}, but it failed to take the advantage of the inner SSMs matrix and still used custom normalization layers like AdaLN for transferring the style. In this context, Mamba was utilized not for merging content and style directly, but primarily as a feature extractor to exploit its fast inference speed. Notably, the combined AdaLN+Mamba model exhibited limited generalization capabilities, requiring separate training for each new (text, image) pair. Consequently, Mamba facilitated more rapid interaction between image patches, while the style fusion process was primarily handled by the AdaLN layers. On the contrary, we adapt the inner equation of Mamba and provide a full Mamba-based architecture which is able to merge content and style without any other module like AdaLN or cross-attention. Moreover, our method learns how to transfer style and, once trained, can be used with any image-style pair without the need to be retrained.

\section{Method}
\begin{figure*}[!ht]
    \centering
    \includegraphics[width=\textwidth]{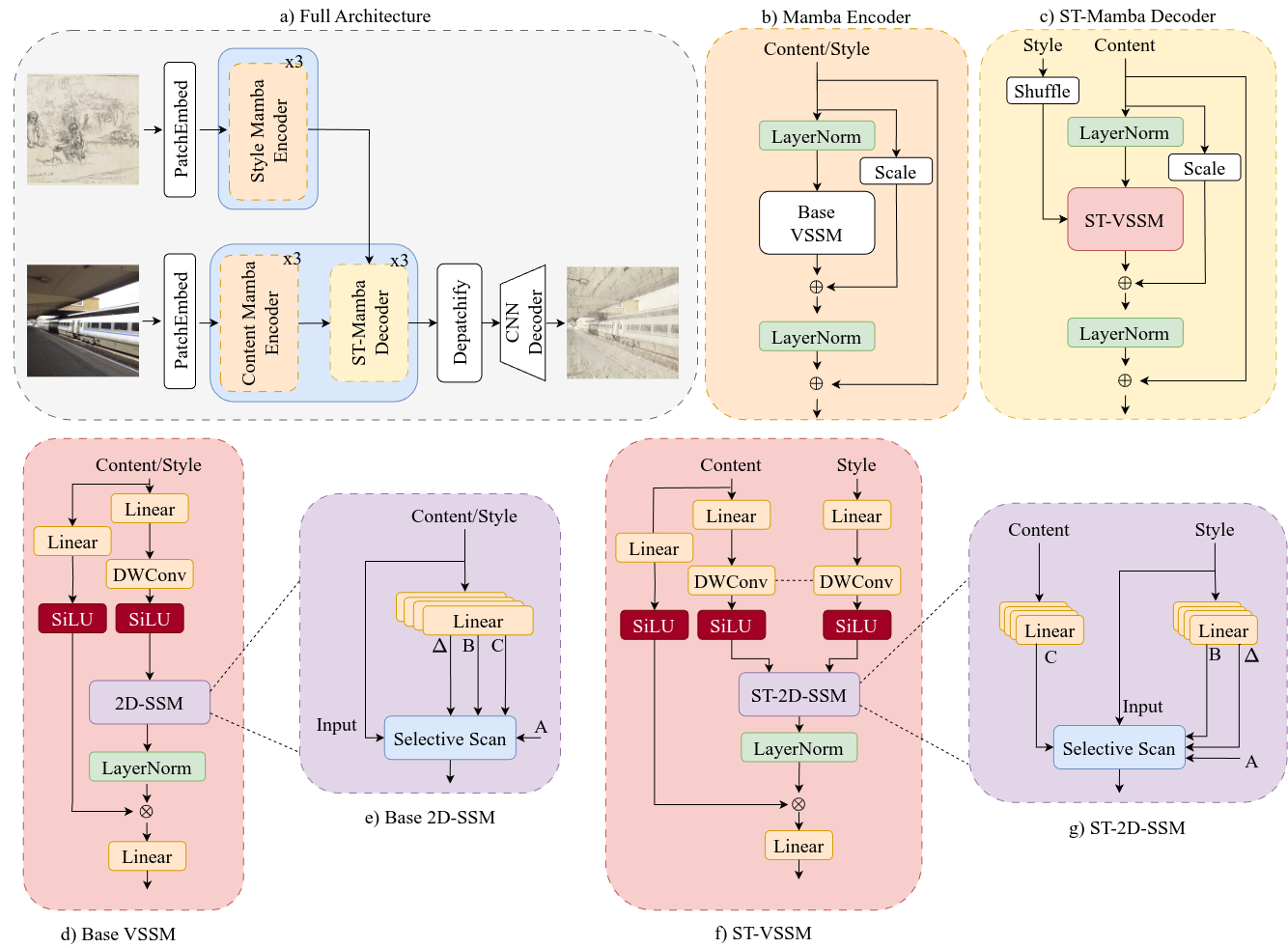}

    \caption{\textit{a}) Mamba-ST full architecture. It takes as input a content and a style image and generates the content image stylized as the style image. \textit{b}) Mamba encoder derived from \cite{liu2024vmambavisualstatespace} with an additional skip connection (rightmost). \textit{c}) Our Mamba-ST Decoder, which takes both style and content as input. In particular, style embeddings are shuffled before passing to ST-VSSM in order to loose spatial information, maintaining only higher level information. \textit{d}) The inner architecture of the Base VSSM. \textit{e}) The inner architecture of the Base 2D-SSM. \textit{f)} Our ST-VSSM. Notably, DWConv is shared among content and style embedding. \textit{g}) Our modified ST 2D-SSM, where the matrices $A$,$B$ and $\Delta$ are computed from the style, the input of the selective scan are the style embedding and the matrix C is calculated using the content.}
    \label{fig:architecture}
\end{figure*}

In this section, the proposed architecture is introduced by firstly describing Mamba and then showing how its inner equation can be adapted to perform style transfer.

\subsection{Background on Mamba}
State-Space Models, like S4 \cite{gu2022efficiently}, learn to map a 1-D sequence $x(t) \in \mathbb{R} $ to another 1-D sequence $y(t) \in \mathbb{R} $ as output, maintaining an internal state space $h(t) \in \mathbb{R}^N$, where $N$ is the state size. 
Differently from other sequence learning approaches like transformers, SSMs scale linearly w.r.t. the sequence length. SSMs are described by a linear ordinary differential equation (ODE) system:
\begin{equation}
     \begin{split}
     h'(t) &= A h(t) + B x(t)  \\        
     y(t) &= C h(t) + D x(t) 
     \end{split}
     \label{eq:mamba}
\end{equation}

\noindent where the matrices $A \in \mathbb{R}^{N\times N},B\in \mathbb{R}^{N\times 1},C\in \mathbb{R}^{1\times N} $ and $ D\in \mathbb{R}$ are learnable. In order to be usable in deep learning architectures, a discretization phase is applied to the system. Here we decided to use the zero-order holder (ZOH) rule, where $\Delta$ represents the step parameter.
Specifically, by denoting with $\Bar{A},\Bar{B}$ the discretized matrices, written in a RNN form, the equation becomes:
\begin{equation}
     \begin{split}
     h_{k} &= \Bar{A}h_{k-1} + \bar{B}x_k  \\        
     y_{k} &= Ch_k + Dx_k
     \end{split}
     \label{eq:mamba_disc}
\end{equation}
Given their ability to compress the context in a finite state, recurrent networks are more efficient than transformers, yet their main limitation becomes how well the state can compress the context information \cite{mamba} (\textit{e.g.} understanding the most relevant words in a sentence while ignoring the others). For this reasons, Gu \etal introduced Mamba \cite{mamba}, by adding an input dependency on the matrices $B, C$ and $\Delta$:
\begin{equation}
     \begin{split}
     B = \mathrm{Lin_B}(x), C = \mathrm{Lin_C}(x), \Delta = \mathrm{Lin_{\Delta}}(x)
     \end{split}
     \label{eq:mamba_input_dependency}
\end{equation}
with $\mathrm{Lin}_*$ being a linear, fully-connected layer.

This contribution, combined with an efficient selective scan algorithm for temporal coherency, maintains all the computation efficiency but with a state that is able to better memorize and understand the context information.



\subsection{Overall Architecture}
Given the ability of Mamba to better understand contextual information while maintaining the efficiency of a recurrent network, in this work we aim to adapt Mamba to perform style transfer. To this end, we propose Mamba-ST, whose architecture is shown in Figure \ref{fig:architecture} (a). Our system is composed of three main components: \textit{(i)}~two Mamba Encoders that encode content and style images, respectively, \textit{(ii)}~a Mamba Style Transfer (ST) Decoder (MSTD) that fuses together content and style information and \textit{(iii)}~a CNN decoder to rearrange the decoder output back to an image.

Both content and style images are divided into patches and projected into 1D embeddings using a PatchEmbed layer \cite{rw2019timm}. The PatchEmbed takes the images $I \in \mathbb{R}^{C \times H \times W}$ as input, and produces a series of embeddings $t \in \mathbb{R}^{D\times (h\cdot w)}$, where $h = \frac{H}{p}$ and $w = \frac{W}{p}$ (with $p$ the patch size), and $D$ is the hidden dimension of each embedding. 
Then, we employ the two different domain-encoders, which take as input the corresponding embedding set (\textit{i.e.} content and style), to learn the visual representations of the images. 
After that, content and style representations extracted by the Mamba encoders are fed to the Mamba-ST Decoder (see Fig. \ref{fig:architecture} (c)) which is tasked to merge the two streams of information. 
Finally, a depatchify block is used to obtain a feature map of size $D\times h\times w$ that the CNN decoder transforms to obtain the output image of size $C\times H \times W$.

\begin{figure}[!ht]
    \centering
    \includegraphics[width=0.9\linewidth]{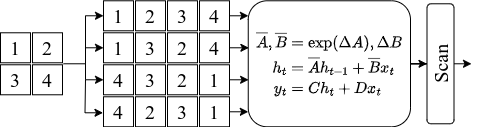}
    \caption{The 2D selective scan with a $2\times 2$ example image.}
    \label{fig:sel_scan}
\end{figure}

\subsection{Mamba Encoder}
Each of the encoders is composed of three Mamba Encoder layers, illustrated in Fig.~\ref{fig:architecture} (b), whose structure is derived from VMamba \cite{liu2024vmambavisualstatespace}, except for a skip connection that we added between each layer, in order to avoid vanishing gradient problem.
After an initial layer normalization, the embeddings are fed to the Base Visual SSM (Base VSSM), illustrated in Fig.~\ref{fig:architecture} (d). 
The Base VSSM structure is achieved by substituting the S6 module employed by Mamba to perform the selective scan, with its 2D counterpart called 2D-SSM (Fig \ref{fig:architecture} (e)).


The 2D-SSM learns the matrices $A,B,C$ and $\Delta$ and models the state of the layer and the output, which is the encoded visual representation of the image. 
The core of 2D-SSM is the selective scan mechanism~\cite{mamba} adapted for 2D sequences (Fig.~\ref{fig:sel_scan}). We follow \cite{liu2024vmambavisualstatespace} and use four different scan directions to maintain spatial information. Then, for each scan, we calculate state and output following Eq.~\eqref{eq:mamba_disc} and merge them together by reordering and summing them. The algorithm is presented in Alg.~\ref{alg:encoder}.

\subsection{Mamba-ST Decoder}
\label{sec:MSTD}
The Mamba-ST Decoder (MSTD) is tasked to merge the content and the style visual representations, extracted by the two Mamba encoder blocks, in a single representation, effectively performing style transfer. Its overall structure is similar to the encoder one, but differs in two main characteristics. First, it takes as input both style and content embeddings, which are fed to a modified version of the base VSSM called ST-VSSM (see Fig.~\ref{fig:architecture}~(f)). Secondly, inside the ST-VSSM, the 2D-SSM is replaced by the proposed ST-2D-SSM (see Fig.~\ref{fig:architecture}~(g)) which is specifically designed to fuse content and style information.

Recently, \cite{ali2024hidden} and \cite{dao2024transformers} showed a duality between Mamba equation and transformer self-attention. In particular, they suggested that query $Q$, key $K$ and value $V$ matrices employed in the self attention equation ($Att = \mathrm{Softmax}(\frac{Q \cdot K^T}{\sqrt{d}}) \cdot V$) can be expressed as: 
\begin{equation}
 Q \approx C, K \approx B, V \approx X   
\end{equation}
\noindent where $X$ is the input sequence. Following this symmetry between Mamba and self-attention, our intuition was to mimic a cross-attention mechanism by letting $A,B$ and $\Delta$ matrices be dependent from the style source $s$, while making $C$ dependent from the content $x$ as follows:
\begin{equation}
\begin{split}
 B = \mathrm{Lin_B}(s), \Delta = \mathrm{Lin_{\Delta}}(s), C = \mathrm{Lin_C}(x)
 \end{split}
 \label{eq:mamba_input_style_dependency}  
\end{equation}
Similarly, we decided to pass the encoded style features instead of the content as input sequence for the ST-2D-SSM. More in detail, based on Eq.~\eqref{eq:mamba_disc}, we incorporate the style information inside the state, as the equation for the state depends only on $A$ and $B$ matrices:
\begin{equation}
     h_{k} = \Bar{A}h_{k-1} + \bar{B}s_k     
\end{equation}
Then, the output is made dependent on both style and content since $C$ is derived from the content image:
\begin{equation}
     y_{k} = Ch_k  
\end{equation}
This allows to effectively merge content and style information and to perform style transfer.

The inner selective-scan mechanism inside the decoder layer is the same as the one inside the encoder. Furthermore, in order to remove content details from the style that could jeopardize the style transfer, we decided to apply a random shuffle to the style embedding, as shown in Fig.~\ref{fig:architecture}~(c). In this way, the hidden state of the model loses every information about the content of the style picture, leaving only style information. In Alg.~\ref{alg:decoder}, we provide the full algorithm description of the decoder block.

\subsection{Losses}
\label{sec:losses}
We train our model using two perceptual losses. The content loss $\mathcal{L}_C$ focuses on preserving the content of the original image, while the style loss $\mathcal{L}_S$ aims to transfer the style of the source image to the target image.

We implement the two losses using a pretrained VGG19 model following \cite{huang2017arbitrary, an2021artflow, deng2022stytr2}. Let $x_c$ be the content image, $x_s$ be the style image and $x_g$ be the generated image. Given $N_l$ the number of the layers selected from the VGG19, we define $\phi_i(x)$ as the features extracted from the $i$-th layer with as input the image $x$. The content loss is defined as:
\begin{equation}
    \mathcal{L}_C=\frac{1}{N_l}\sum_{i=0}^{N_l}\left\lVert\phi_i\left(x_g\right)-\phi_i\left(x_c\right)\right\rVert_2
\end{equation}
The style loss is instead defined as:
\begin{equation}
    \begin{split}
        \mathcal{L}_S=\frac{1}{N_l}\sum_{i=0}^{N_l}(\lVert\mu\left(\phi_i\left(x_g\right)\right)-\mu\left(\phi_i\left(x_s\right)\right)\rVert_2 \\
        +\lVert\sigma\left(\phi_i\left(x_g\right)\right)-\sigma\left(\phi_i\left(x_s\right)\right)\rVert_2)
    \end{split}
\end{equation}
where $\mu(\cdot)$ is the mean of a given feature map and $\sigma(\cdot)$ is the standard deviation of a given feature map. 

Furthermore, we also use two identity losses\cite{park2019arbitrary, deng2022stytr2}. These help in learning better representations for both content and style. Let $x_g^c$ be the image generated using the content image $x_c$ as both style and content information, and $x_g^s$ be the generated image using the style image $x_s$ as both style and content information, we then define:
\begin{align}    
    \mathcal{L}_{id1} &=||x_g^c-x_c||_2+||x_g^s-x_s||_2 ,\\
    \begin{split}
    \mathcal{L}_{id2}&=\frac{1}{N_l}\sum_{i=0}^{N_l}(||\phi(x_g^c)-\phi(x_c)||_2\\
    &\qquad+||\phi(x_g^s)-\phi(x_s)||_2)
    \end{split}
\end{align}
The final loss which we use to train our model is:
\begin{equation}
    \mathcal{L}=\lambda_C\mathcal{L}_C+\lambda_S\mathcal{L}_S+\lambda_{id1}\mathcal{L}_{id1}+\lambda_{id2}\mathcal{L}_{id2},
\end{equation}
with $\lambda_C=7,\lambda_S=10,\lambda_{id1}=70$ and $\lambda_{id2}=1$ in order to balance the magnitude of each loss \cite{deng2022stytr2}.

\section{Experiments}

\begin{figure*}[!ht]
    \centering
    \includegraphics[width=\linewidth]{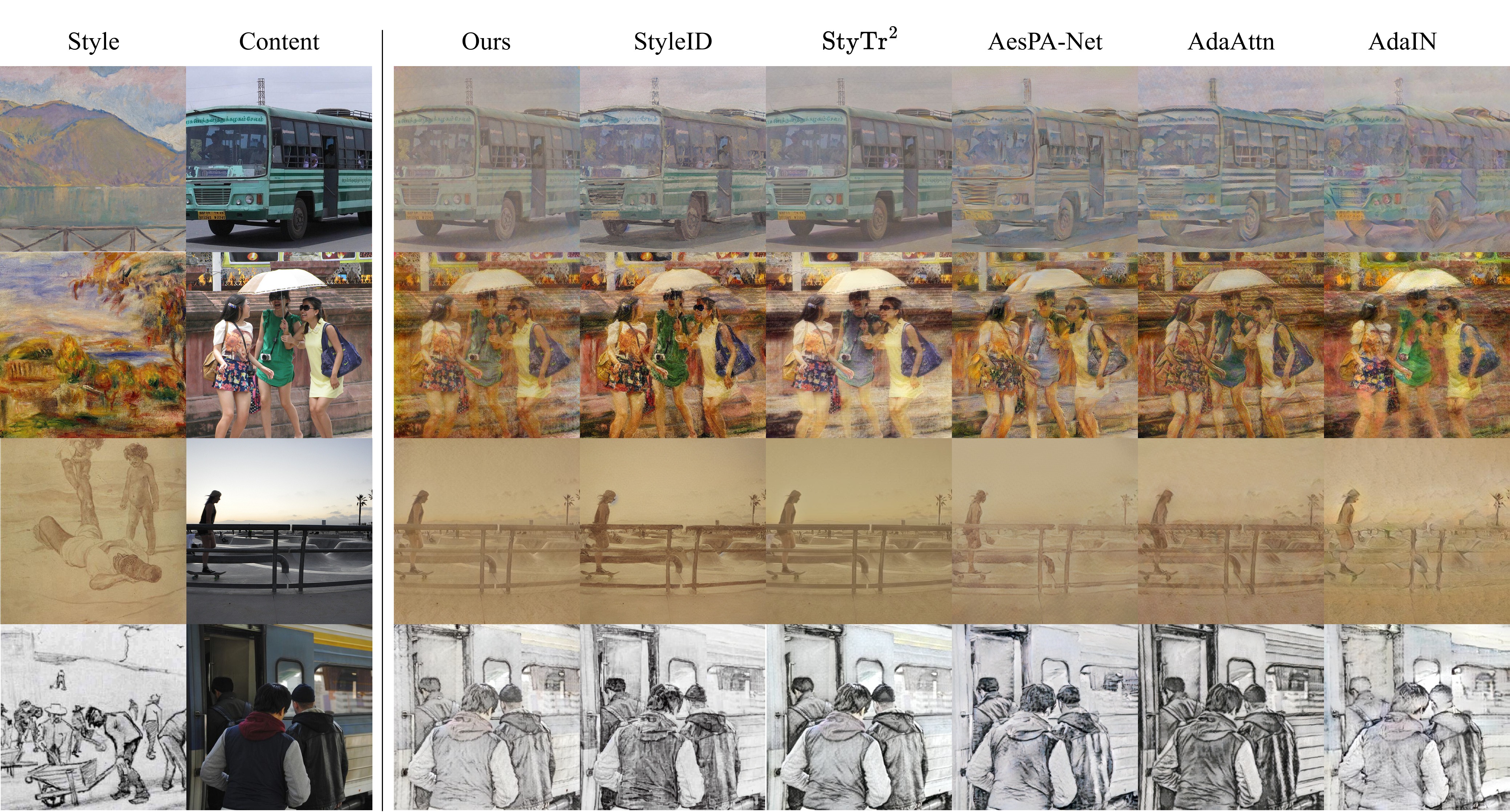}
    \caption{Visual comparison with the current state-of-the-art models.}
    \label{fig:comparison}
\end{figure*}

\paragraph{Implementation details}
We use the COCO dataset \cite{lin2014microsoft} as our content dataset and the WikiArt dataset \cite{tan2018improved} as our style dataset to train our model. We adopt the same hyperparameters setting as \cite{deng2022stytr2}, with the exception of the learning rate, which we set to 0.00005 without utilizing any warm-up period. Our model is trained for a total of 160,000 iterations with a batch size of 8 on a single NVIDIA L40S GPU. Moreover, we set $p$ (\textit{i.e.} the patch size) to 8.

\paragraph{Evaluation details}
We compare our model both qualitatively and quantitatively with several state-of-the-art models designed for image-to-image style transfer: StyleID\cite{chung2024style}, AesPA-Net \cite{hong2023aespa}, StyTr$^2$\cite{deng2022stytr2}, AdaAttn\cite{liu2021adaattn} AdaIN\cite{huang2017arbitrary}. 
We intentionally avoid comparing with style transfer methods which use textual descriptions as style condition instead of an image (including StyleMamba\footnote{Furthermore, no code is available for this paper} \cite{wang2024stylemamba}) since it would not be a fair comparison.

To quantitatively evaluate our model, we employ four primary metrics: ArtFID \cite{wright2022artfid}, FID \cite{heusel2017gans}, LPIPS \cite{zhang2018unreasonable}, and CFSD \cite{chung2024style}. We did not consider content loss $\mathcal{L}_C$ and style loss $\mathcal{L}_S$ as complementary evaluation metrics since \cite{chung2024style} noted that utilizing these losses for both training and evaluation would introduce evaluation biases.

More in detail, FID \cite{heusel2017gans} measures the overall similarity between generated and style images distributions. 
LPIPS \cite{zhang2018unreasonable} evaluates the content preservation between a source image and the stylized image, thereby measuring how well the content is preserved in the stylized image. ArtFID \cite{wright2022artfid} is a metric found to be highly correlated with human judgment, and fuses together style transfer and content preservation. ArtFID is calculated as $(1+\mathrm{FID})\cdot(1+\mathrm{LPIPS}).$

CFSD (Content Feature Structural Distance) \cite{chung2024style} is a metric designed to address the limitations of LPIPS \cite{chung2024style, geirhos2018imagenet}. Indeed, LPIPS utilizes a feature space extracted by a pretrained AlexNet \cite{krizhevsky2012imagenet}, which is trained on the ImageNet dataset \cite{deng2009imagenet} for classification tasks. However, ImageNet is known to be texture-biased \cite{geirhos2018imagenet}, meaning that style injection can influence the LPIPS measure, potentially leading to inaccurate assessments of content preservation. CFSD has been hence introduced as distance metric based on the spatial correlation between image patches.

For each model, we calculate the metrics on 800 generated samples obtained by randomly sampling 40 style images and 20 content images  following \cite{chung2024style}. Furthermore, we calculate the inference time (s) to generate the 800 images and memory usage (MebiByte, MiB) with batch size 1. 

\paragraph{Quantitative analysis}
\begin{table*}[!ht]
    \centering
    \begin{tabular}{c?c|c|c|c?c|c}
    \toprule
         & \multicolumn{4}{c?}{Metrics} & \multicolumn{2}{c}{Time and Memory usage} \\
         \midrule
        Model & ArtFID $\downarrow$ & FID $\downarrow$ & LPIPS $\downarrow$ & CFSD $\downarrow$ & Time (s) $\downarrow$ & Memory Usage (MiB) $\downarrow$\\ 
        \midrule
        AdaIN \cite{huang2017arbitrary} & \textcolor{red}{27.81} & \textcolor{red}{16.80} & 0.56 & 0.35 & 12.26 & 824 \\ 
        AdaAttN \cite{liu2021adaattn} & 30.81 & 19.46 & 0.51 &  \textcolor{red}{0.32} & 23.89 & 5554 \\
        StyTr$^2$ \cite{deng2022stytr2} & 29.31 & 18.77 & \textbf{0.48} & \textcolor{red}{0.32} & 52.35 & 2160 \\ 
        AesPA-Net \cite{hong2023aespa} & 35.45 & 22.85 & \textcolor{red}{0.49} & 0.33  & 165.99 & 4184 \\ 
        StyleID \cite{chung2024style}  & 28.65 & 18.29 & \textcolor{red}{0.49} & \textbf{0.29} & 2744.38 & 19930 \\ 
        \midrule
        \textbf{Ours} & \textbf{27.11} & \textbf{16.75} & 0.53 & 0.33 & 24.70 & 1414 \\ 
        \bottomrule

    \end{tabular}
    \caption{Performance comparison of the SOTA models and our proposed model on $512 \times 512$ resolution. The best result for each metric is highlighted in bold, while the second-best result is marked in red. Time is calculated generating the entire 800 stylized images. Memory usage is calculated with batch 1.}
    \label{tab:results}
\end{table*}
\begin{table*}
    \centering
    \begin{tabular}{c?c|c|c|c?c|c}
    \toprule
         & \multicolumn{4}{c?}{Metrics} & \multicolumn{2}{c}{Time and Memory usage} \\
         \midrule
         & ArtFID $\downarrow$ & FID $\downarrow$ & LPIPS $\downarrow$ &  CFSD $\downarrow$ & Time (s) $\downarrow$ & Memory Usage (MiB) $\downarrow$\\ 
        \midrule
        \makecell{Content as input 2D-SSM} & 28.17 & 17.76 & \textcolor{red}{0.50} & \textbf{0.33} & 24.71 & 1414 \\
        \midrule
        \# Scan Directions = 1 & 34.69 & 21.49 & 0.54 & 0.63 & 15.84 & 1288  \\
        \# Scan Directions = 2 & 28.17 & 17.91 & \textbf{0.49} & \textbf{0.33} & 19.05 & 1294 \\
        \midrule
        Dim state=8  & \textcolor{red}{27.26} & 16.79 & 0.53 & \textbf{0.33} & 22.93 &  1424 \\
        Dim state=32 & 28.34 & 17.33 & 0.55 & 0.34 & 27.73 & 1446 \\
        \midrule
        \makecell{w/o random in inference} & 27.63 & \textbf{16.25} & 0.60 & \textcolor{red}{0.36} & 24.39 & 1430 \\
        \midrule
        \textbf{Ours} & \textbf{27.11} & \textcolor{red}{16.75} & 0.53 & \textbf{0.33} & 24.70 & 1414 \\ 
        \bottomrule
    \end{tabular}
    \caption{Various ablation studies. The best result for each metric is highlighted in bold, while the second-best result is marked in red. Time is calculated generating the entire 800 stylized images. Memory usage is calculated with batch 1.}
    \label{tab:ablation}
\end{table*}

In Tab.~\ref{tab:results} quantitative evaluation of several state-of-the-art models w.r.t.~the proposed system is presented. Notably, our method outperforms previous architecture in ArtFID, which, as previously discussed, is strongly correlated with human judgment. Additionally, our method achieves the lowest FID, indicating superior style transfer to the content image. On the other side, in terms of content preservation metrics (\textit{i.e.}~LPIPS and CFSD) our model has comparable, albeit slight worse performance. This trade-off highlights our model capability to transfer the style at the cost of a marginal reduction of content preservation. 
However, we argue that, compared to the best SOTA model (AdaIN) in terms of FID and ArtFID we are able to also improve LPIPS and CFSD and, if compared with the best SOTA model (StyleID) in terms of LPIPS or CFSD, we are able to improve both FID and ArtFID.

AdaIN, thanks to its lightweight backbone, is also the most efficient model in terms of both time and memory usage. Specifically, it can generate 800 images (derived from 20 content images combined with 40 style images) in just 12.26 seconds, while maintaining minimal memory consumption. At the same time though, even if it surprisingly achieves the second best results in ArtFID and FID, it heavily falls behind (it is indeed the worst) in terms of LPIPS and CFSD meaning that it fails in preserving the correct content in the samples. The second fastest model is AdaAttn; however, it demands significantly higher memory capacity. Finally, our proposed model strikes a balance between time and memory usage, requiring low memory while delivering an acceptable inference time. This is particularly evident when comparing with diffusion-based models like StyleID, which is both memory-intensive and time-consuming.

\paragraph{Qualitative analysis}
Qualitative comparisons are shown in Fig.~\ref{fig:comparison}. As it can be seen, we are able to achieve comparable results w.r.t. the current state-of-the-art models. Looking at the figure, AdaIN is able to correctly apply the style, but the overall content is greatly altered as LPIPS and CFSD value in Tab.~\ref{tab:results} already showed. On the other side, StyTr$^2$, AesPA-Net and AdaAttn sometimes struggle to maintain color coherency when applying the style. For example, in the second row the middle girl's dress is turned to blue instead of green which was the color in the original content image. Finally, StyleID is able to produce coherent images that present both the original content and the style features, but with very high contrast and saturation in the colors which is typical of diffusion models. This may alter the faithful application of styles characterized by soft colors (see first and third rows) or, on the other side, push too much the application of high-contrast styles (see second row). The fourth row instead provides examples where every method is capable of producing satisfactory results. Finally, our method represents a trade-off between the previous style transfer models. We are able to both apply styles while maintaining the correct color coherency of the content image, but, at the same time, without excessively changing the saturation and contrast of the generated samples.
\begin{figure}[!ht]
   \centering
   \includegraphics[width=0.9\linewidth]{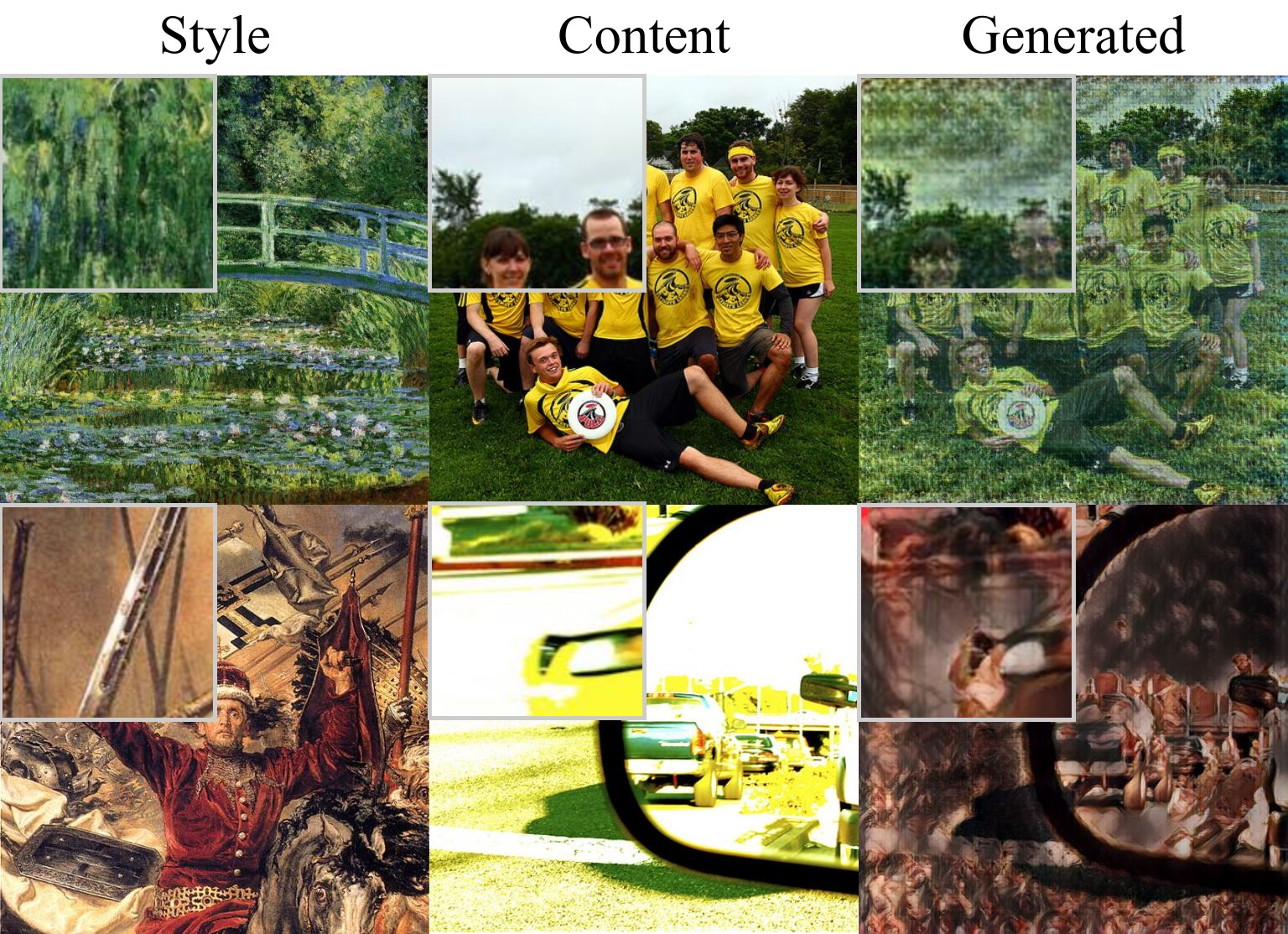}
   \caption{Zoomed results which show the patch problem inside the results. Gaps are present between each patch in the results and the model failed to uniformly apply the style.}
   \label{fig:limitation}
\end{figure}

Despite the excellent results, sometimes the model fails to correctly apply the style, as shown in Fig.~\ref{fig:limitation}. Sometimes it reproduces non-homogeneous patches inside the output images with a gap between them. A possible reason is that Mamba-ST inherits RNNs limitations; in some cases, it is difficult to ensure continuity between the patches due to the memorization of context information inside the state.

\paragraph{Ablation}

\begin{figure}[!ht]
    \centering
    \includegraphics[width=0.9\linewidth]{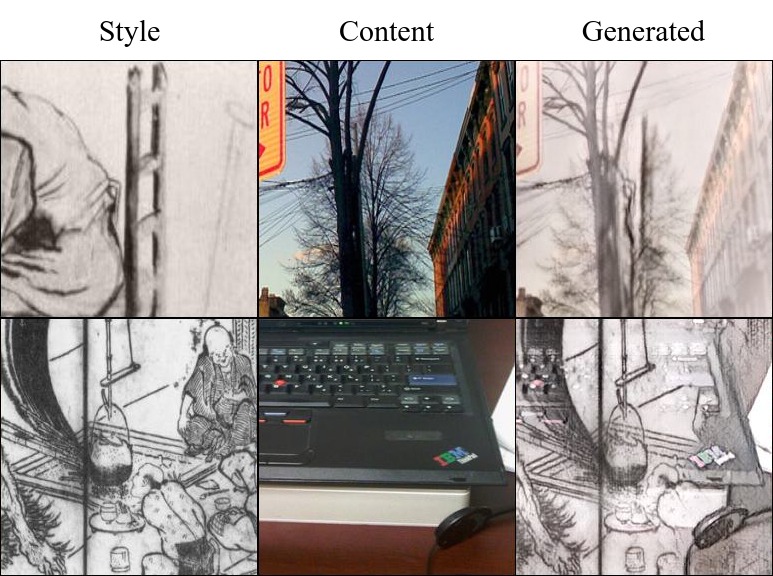}
    \caption{Ablation study of our model without the random shuffle in the style embedding. The generated images become a blend of both context and style images.}
    \label{fig:ablation_rnd}
\end{figure}

Finally, we performed several ablation studies to determine the optimal configuration for our system. Results are shown in Tab.~\ref{tab:ablation}. First, we investigated the effect of passing the content image instead of style to the selective scan. This resulted in overall worse performance. Also looking at the mathematical analysis provided in the supplementary Sec.~\ref{sec:proof}, we ultimately opted for passing the style as input to the selective scan. The second and third rows of Tab.~\ref{tab:ablation} show the performance of our model when trained using 1 or 2 scan directions instead of~4. When adopting a single direction, the overall performance drop largely while 2 directions improve a bit, yet still being lower than the final architecture with 4 directions. The inference time with 4 scans is only slightly worse. In the fourth and fifth rows of Tab.~\ref{tab:ablation} we show the effect of varying the dimension of the Mamba internal state. The performance improves when increasing the dimension from 8 to 16 (our final model), but drop when further increasing it from 16 to 32. Based on these findings, we select 16 as the dimension of the internal state. Finally, in the sixth row of the table, we tested removing the shuffling module at inference. The improved FID indicates that the model more accurately captures the stylistic features of the reference image, leading to good stylized images. Nevertheless, this adjustment led to a large deterioration in LPIPS. This outcome suggests that the model might overly emphasize the style at the expense of preserving the content fidelity. 
Without shuffling, the spatial and structural information within the style image is maintained, ultimately leading to a blended version of the content and style. The shuffling module is thus necessary even at inference for effectively capturing high-level style information.

In conclusion, Fig.~\ref{fig:ablation_rnd} presents examples generated by our best model when shuffling is disabled during the training phase too. The absence of shuffling prevents the model from isolating style information in the generated output. Instead, it retains content details from the style image as well. Consequently, the output is a blended version of both the content and style images, rather than a proper transfer of style.


%

\section{Conclusion}
In this work we investigated the adaptation of Mamba inner equation for image driven style transfer, leveraging its lightweight capability to lower time consumption and memory usage w.r.t. the state of the art. Specifically, we propose a new Mamba block, called Mamba-ST Decoder which is able to accept two streams of information as input and fuse them together in a single output. Finally, we provide an extensive set of comparison with SOTA models and a comprehensive set of ablation studies proving the efficacy of the proposed solution.

\section{Acknowledgments}
This work was funded by ``Partenariato FAIR (Future Artificial Intelligence Research) - PE00000013, CUP J33C22002830006" funded by the European Union - NextGenerationEU through the italian MUR within NRRP.


{\small
\bibliographystyle{ieee_fullname}
\bibliography{egbib}

\begin{thebibliography}{10}\itemsep=-1pt

\bibitem{ali2024hidden}
Ameen Ali, Itamar Zimerman, and Lior Wolf.
\newblock The hidden attention of mamba models.
\newblock {\em arXiv preprint arXiv:2403.01590}, 2024.

\bibitem{an2021artflow}
Jie An, Siyu Huang, Yibing Song, Dejing Dou, Wei Liu, and Jiebo Luo.
\newblock Artflow: Unbiased image style transfer via reversible neural flows.
\newblock In {\em Proceedings of the IEEE/CVF Conference on Computer Vision and Pattern Recognition}, pages 862--871, 2021.

\bibitem{carion2020end}
Nicolas Carion, Francisco Massa, Gabriel Synnaeve, Nicolas Usunier, Alexander Kirillov, and Sergey Zagoruyko.
\newblock End-to-end object detection with transformers.
\newblock In {\em European conference on computer vision}, pages 213--229. Springer, 2020.

\bibitem{chandran2021adaptive}
Prashanth Chandran, Gaspard Zoss, Paulo Gotardo, Markus Gross, and Derek Bradley.
\newblock Adaptive convolutions for structure-aware style transfer.
\newblock In {\em Proceedings of the IEEE/CVF conference on computer vision and pattern recognition}, pages 7972--7981, 2021.

\bibitem{choromanski2020rethinking}
Krzysztof Choromanski, Valerii Likhosherstov, David Dohan, Xingyou Song, Andreea Gane, Tamas Sarlos, Peter Hawkins, Jared Davis, Afroz Mohiuddin, Lukasz Kaiser, et~al.
\newblock Rethinking attention with performers.
\newblock {\em arXiv preprint arXiv:2009.14794}, 2020.

\bibitem{chung2024style}
Jiwoo Chung, Sangeek Hyun, and Jae-Pil Heo.
\newblock Style injection in diffusion: A training-free approach for adapting large-scale diffusion models for style transfer.
\newblock In {\em Proceedings of the IEEE/CVF Conference on Computer Vision and Pattern Recognition}, pages 8795--8805, 2024.

\bibitem{dao2023flashattention}
Tri Dao.
\newblock Flashattention-2: Faster attention with better parallelism and work partitioning.
\newblock {\em arXiv preprint arXiv:2307.08691}, 2023.

\bibitem{dao2022flashattention}
Tri Dao, Dan Fu, Stefano Ermon, Atri Rudra, and Christopher R{\'e}.
\newblock Flashattention: Fast and memory-efficient exact attention with io-awareness.
\newblock {\em Advances in Neural Information Processing Systems}, 35:16344--16359, 2022.

\bibitem{dao2024transformers}
Tri Dao and Albert Gu.
\newblock {Transformers are SSMs: Generalized models and efficient algorithms through structured state space duality}.
\newblock {\em arXiv preprint arXiv:2405.21060}, 2024.

\bibitem{deng2009imagenet}
Jia Deng, Wei Dong, Richard Socher, Li-Jia Li, Kai Li, and Li Fei-Fei.
\newblock Imagenet: A large-scale hierarchical image database.
\newblock In {\em 2009 IEEE conference on computer vision and pattern recognition}, pages 248--255. Ieee, 2009.

\bibitem{deng2022stytr2}
Yingying Deng, Fan Tang, Weiming Dong, Chongyang Ma, Xingjia Pan, Lei Wang, and Changsheng Xu.
\newblock Stytr2: Image style transfer with transformers.
\newblock In {\em Proceedings of the IEEE/CVF conference on computer vision and pattern recognition}, pages 11326--11336, 2022.

\bibitem{deng2020arbitrary}
Yingying Deng, Fan Tang, Weiming Dong, Wen Sun, Feiyue Huang, and Changsheng Xu.
\newblock Arbitrary style transfer via multi-adaptation network.
\newblock In {\em Proceedings of the 28th ACM international conference on multimedia}, pages 2719--2727, 2020.

\bibitem{ding2023longnet}
Jiayu Ding, Shuming Ma, Li Dong, Xingxing Zhang, Shaohan Huang, Wenhui Wang, Nanning Zheng, and Furu Wei.
\newblock Longnet: Scaling transformers to 1,000,000,000 tokens.
\newblock {\em arXiv preprint arXiv:2307.02486}, 2023.

\bibitem{DBLP:journals/corr/abs-2010-11929}
Alexey Dosovitskiy, Lucas Beyer, Alexander Kolesnikov, Dirk Weissenborn, Xiaohua Zhai, Thomas Unterthiner, Mostafa Dehghani, Matthias Minderer, Georg Heigold, Sylvain Gelly, Jakob Uszkoreit, and Neil Houlsby.
\newblock An image is worth 16x16 words: Transformers for image recognition at scale.
\newblock {\em CoRR}, abs/2010.11929, 2020.

\bibitem{gatys2016image}
Leon~A Gatys, Alexander~S Ecker, and Matthias Bethge.
\newblock Image style transfer using convolutional neural networks.
\newblock In {\em Proceedings of the IEEE conference on computer vision and pattern recognition}, pages 2414--2423, 2016.

\bibitem{geirhos2018imagenet}
Robert Geirhos, Patricia Rubisch, Claudio Michaelis, Matthias Bethge, Felix~A Wichmann, and Wieland Brendel.
\newblock Imagenet-trained cnns are biased towards texture; increasing shape bias improves accuracy and robustness.
\newblock {\em arXiv preprint arXiv:1811.12231}, 2018.

\bibitem{mamba}
Albert Gu and Tri Dao.
\newblock Mamba: Linear-time sequence modeling with selective state spaces.
\newblock {\em arXiv preprint arXiv:2312.00752}, 2023.

\bibitem{gu2022efficiently}
Albert Gu, Karan Goel, and Christopher R\'e.
\newblock Efficiently modeling long sequences with structured state spaces.
\newblock In {\em The International Conference on Learning Representations ({ICLR})}, 2022.

\bibitem{gu2021combining}
Albert Gu, Isys Johnson, Karan Goel, Khaled Saab, Tri Dao, Atri Rudra, and Christopher R{\'e}.
\newblock Combining recurrent, convolutional, and continuous-time models with linear state space layers.
\newblock {\em Advances in neural information processing systems}, 34:572--585, 2021.

\bibitem{guo2024mambair}
Hang Guo, Jinmin Li, Tao Dai, Zhihao Ouyang, Xudong Ren, and Shu-Tao Xia.
\newblock Mambair: A simple baseline for image restoration with state-space model.
\newblock {\em arXiv preprint arXiv:2402.15648}, 2024.

\bibitem{heusel2017gans}
Martin Heusel, Hubert Ramsauer, Thomas Unterthiner, Bernhard Nessler, and Sepp Hochreiter.
\newblock Gans trained by a two time-scale update rule converge to a local nash equilibrium.
\newblock {\em Advances in neural information processing systems}, 30, 2017.

\bibitem{hong2023aespa}
Kibeom Hong, Seogkyu Jeon, Junsoo Lee, Namhyuk Ahn, Kunhee Kim, Pilhyeon Lee, Daesik Kim, Youngjung Uh, and Hyeran Byun.
\newblock Aespa-net: Aesthetic pattern-aware style transfer networks.
\newblock In {\em Proceedings of the IEEE/CVF International Conference on Computer Vision}, pages 22758--22767, 2023.

\bibitem{hu2024zigma}
Vincent~Tao Hu, Stefan~Andreas Baumann, Ming Gui, Olga Grebenkova, Pingchuan Ma, Johannes Fischer, and Bjorn Ommer.
\newblock Zigma: Zigzag mamba diffusion model.
\newblock {\em arXiv preprint arXiv:2403.13802}, 2024.

\bibitem{huang2017arbitrary}
Xun Huang and Serge Belongie.
\newblock Arbitrary style transfer in real-time with adaptive instance normalization.
\newblock In {\em Proceedings of the IEEE international conference on computer vision}, pages 1501--1510, 2017.

\bibitem{kim2022diffusionclip}
Gwanghyun Kim, Taesung Kwon, and Jong~Chul Ye.
\newblock Diffusionclip: Text-guided diffusion models for robust image manipulation.
\newblock In {\em Proceedings of the IEEE/CVF conference on computer vision and pattern recognition}, pages 2426--2435, 2022.

\bibitem{krizhevsky2012imagenet}
Alex Krizhevsky, Ilya Sutskever, and Geoffrey~E Hinton.
\newblock Imagenet classification with deep convolutional neural networks.
\newblock {\em Advances in neural information processing systems}, 25, 2012.

\bibitem{kwon2022clipstyler}
Gihyun Kwon and Jong~Chul Ye.
\newblock Clipstyler: Image style transfer with a single text condition.
\newblock In {\em Proceedings of the IEEE/CVF Conference on Computer Vision and Pattern Recognition}, pages 18062--18071, 2022.

\bibitem{kwon2022diffusion}
Gihyun Kwon and Jong~Chul Ye.
\newblock Diffusion-based image translation using disentangled style and content representation.
\newblock {\em arXiv preprint arXiv:2209.15264}, 2022.

\bibitem{li2017universal}
Yijun Li, Chen Fang, Jimei Yang, Zhaowen Wang, Xin Lu, and Ming-Hsuan Yang.
\newblock Universal style transfer via feature transforms.
\newblock {\em Advances in neural information processing systems}, 30, 2017.

\bibitem{lin2014microsoft}
Tsung-Yi Lin, Michael Maire, Serge Belongie, James Hays, Pietro Perona, Deva Ramanan, Piotr Doll{\'a}r, and C~Lawrence Zitnick.
\newblock Microsoft coco: Common objects in context.
\newblock In {\em Computer Vision--ECCV 2014: 13th European Conference, Zurich, Switzerland, September 6-12, 2014, Proceedings, Part V 13}, pages 740--755. Springer, 2014.

\bibitem{liu2021adaattn}
Songhua Liu, Tianwei Lin, Dongliang He, Fu Li, Meiling Wang, Xin Li, Zhengxing Sun, Qian Li, and Errui Ding.
\newblock Adaattn: Revisit attention mechanism in arbitrary neural style transfer.
\newblock In {\em Proceedings of the IEEE/CVF international conference on computer vision}, pages 6649--6658, 2021.

\bibitem{liu2024vmambavisualstatespace}
Yue Liu, Yunjie Tian, Yuzhong Zhao, Hongtian Yu, Lingxi Xie, Yaowei Wang, Qixiang Ye, and Yunfan Liu.
\newblock Vmamba: Visual state space model, 2024.

\bibitem{lu2019closed}
Ming Lu, Hao Zhao, Anbang Yao, Yurong Chen, Feng Xu, and Li Zhang.
\newblock A closed-form solution to universal style transfer.
\newblock In {\em Proceedings of the IEEE/CVF International Conference on Computer Vision}, pages 5952--5961, 2019.

\bibitem{park2019arbitrary}
Dae~Young Park and Kwang~Hee Lee.
\newblock Arbitrary style transfer with style-attentional networks.
\newblock In {\em proceedings of the IEEE/CVF conference on computer vision and pattern recognition}, pages 5880--5888, 2019.

\bibitem{rombach2022high}
Robin Rombach, Andreas Blattmann, Dominik Lorenz, Patrick Esser, and Bj{\"o}rn Ommer.
\newblock High-resolution image synthesis with latent diffusion models.
\newblock In {\em Proceedings of the IEEE/CVF conference on computer vision and pattern recognition}, pages 10684--10695, 2022.

\bibitem{shi2024vmambair}
Yuan Shi, Bin Xia, Xiaoyu Jin, Xing Wang, Tianyu Zhao, Xin Xia, Xuefeng Xiao, and Wenming Yang.
\newblock Vmambair: Visual state space model for image restoration.
\newblock {\em arXiv preprint arXiv:2403.11423}, 2024.

\bibitem{sun2019videobert}
Chen Sun, Austin Myers, Carl Vondrick, Kevin Murphy, and Cordelia Schmid.
\newblock Videobert: A joint model for video and language representation learning.
\newblock In {\em Proceedings of the IEEE/CVF international conference on computer vision}, pages 7464--7473, 2019.

\bibitem{tan2018improved}
Wei~Ren Tan, Chee~Seng Chan, Hernan~E Aguirre, and Kiyoshi Tanaka.
\newblock Improved artgan for conditional synthesis of natural image and artwork.
\newblock {\em IEEE Transactions on Image Processing}, 28(1):394--409, 2018.

\bibitem{wang2024mambabyte}
Junxiong Wang, Tushaar Gangavarapu, Jing~Nathan Yan, and Alexander~M Rush.
\newblock Mambabyte: Token-free selective state space model.
\newblock {\em arXiv preprint arXiv:2401.13660}, 2024.

\bibitem{wang2020linformer}
Sinong Wang, Belinda~Z Li, Madian Khabsa, Han Fang, and Hao Ma.
\newblock Linformer: Self-attention with linear complexity.
\newblock {\em arXiv preprint arXiv:2006.04768}, 2020.

\bibitem{wang2024stylemamba}
Zijia Wang and Zhi-Song Liu.
\newblock Stylemamba: State space model for efficient text-driven image style transfer.
\newblock {\em arXiv preprint arXiv:2405.05027}, 2024.

\bibitem{wang2020diversified}
Zhizhong Wang, Lei Zhao, Haibo Chen, Lihong Qiu, Qihang Mo, Sihuan Lin, Wei Xing, and Dongming Lu.
\newblock Diversified arbitrary style transfer via deep feature perturbation.
\newblock In {\em Proceedings of the IEEE/CVF Conference on Computer Vision and Pattern Recognition}, pages 7789--7798, 2020.

\bibitem{wang2023stylediffusion}
Zhizhong Wang, Lei Zhao, and Wei Xing.
\newblock Stylediffusion: Controllable disentangled style transfer via diffusion models.
\newblock In {\em Proceedings of the IEEE/CVF International Conference on Computer Vision}, pages 7677--7689, 2023.

\bibitem{rw2019timm}
Ross Wightman.
\newblock Pytorch image models.
\newblock \url{https://github.com/rwightman/pytorch-image-models}, 2019.

\bibitem{wright2022artfid}
Matthias Wright and Bj{\"o}rn Ommer.
\newblock Artfid: Quantitative evaluation of neural style transfer.
\newblock In {\em DAGM German Conference on Pattern Recognition}, pages 560--576. Springer, 2022.

\bibitem{yang2020learning}
Fuzhi Yang, Huan Yang, Jianlong Fu, Hongtao Lu, and Baining Guo.
\newblock Learning texture transformer network for image super-resolution.
\newblock In {\em Proceedings of the IEEE/CVF conference on computer vision and pattern recognition}, pages 5791--5800, 2020.

\bibitem{ye2019cross}
Linwei Ye, Mrigank Rochan, Zhi Liu, and Yang Wang.
\newblock Cross-modal self-attention network for referring image segmentation.
\newblock In {\em Proceedings of the IEEE/CVF conference on computer vision and pattern recognition}, pages 10502--10511, 2019.

\bibitem{zhang2018unreasonable}
Richard Zhang, Phillip Isola, Alexei~A Efros, Eli Shechtman, and Oliver Wang.
\newblock The unreasonable effectiveness of deep features as a perceptual metric.
\newblock In {\em Proceedings of the IEEE conference on computer vision and pattern recognition}, pages 586--595, 2018.

\bibitem{zhang2023inversion}
Yuxin Zhang, Nisha Huang, Fan Tang, Haibin Huang, Chongyang Ma, Weiming Dong, and Changsheng Xu.
\newblock Inversion-based style transfer with diffusion models.
\newblock In {\em Proceedings of the IEEE/CVF conference on computer vision and pattern recognition}, pages 10146--10156, 2023.

\bibitem{zhu2024vision}
Lianghui Zhu, Bencheng Liao, Qian Zhang, Xinlong Wang, Wenyu Liu, and Xinggang Wang.
\newblock Vision mamba: Efficient visual representation learning with bidirectional state space model.
\newblock {\em arXiv preprint arXiv:2401.09417}, 2024.

\end{thebibliography}
}

\newpage

\section{Supplementary Material}

\newcounter{tmp}
\setcounter{tmp}{\value{figure}}
\begin{figure*}[!ht]
\begin{minipage}{0.5\textwidth}
\begin{algorithm}[H]
\caption{Base VSSM}
\label{alg:encoder}
\begin{algorithmic}[1]
\Require Patch sequences $p$
\Ensure Encoded patch $p_e$ 
\State $x \gets \mathrm{Lin}(p)$
\State $z \gets \mathrm{Lin}(p)$
\State $x \gets \mathrm{DWConv}(x)$
\State $x \gets \mathrm{SiLU}(x)$
\For{ \emph{d} in \{scan-directions\}}
    \State $B \gets \mathrm{Lin_B}(x)$
    \State $C \gets \mathrm{Lin_C}(x)$
    \State $\Delta \gets \mathrm{Lin_\Delta}(x)$
    \State $\Bar{A} \gets \Delta \otimes \mathrm{parameter}^A$
    \State $\Bar{B} \gets \Delta \otimes B$
    \State $y_d \gets \mathrm{Selective Scan}(\Bar{A},\Bar{B},C)(x)$
\EndFor
\For{ \emph{d} in \{scan-directions\}}
    $y \gets y + y_d$
\EndFor
\State $y \gets \mathrm{LayerNorm}(y)$
\State $y \gets y * \mathrm{SiLU}(z)$
\State $y \gets \mathrm{Lin}(y)$
\end{algorithmic}
\end{algorithm}
\end{minipage}
\hfill
\begin{minipage}{0.5\textwidth}
\begin{algorithm}[H]
\caption{ST-VSSM}
\label{alg:decoder}
\begin{algorithmic}[1]
\Require Content $c$, Shuffled Style $s$
\Ensure Stylized patches $y$ 
\State $x, s \gets \mathrm{Lin}(c)$, $\mathrm{Lin}(s)$
\State $z \gets \mathrm{Lin}(c)$
\State $x, s \gets \mathrm{DWConv}(x)$, $\mathrm{DWConv}(s)$
\State $x, s \gets \mathrm{SiLU}(x)$, $\mathrm{SiLU}(s)$
\For{ \emph{d} in \{scan-directions\}}
    \State $B \gets \mathrm{Lin_B}(s)$
    \State $C \gets \mathrm{Lin_C}(x)$
    \State $\Delta \gets \mathrm{Lin_\Delta}(s)$
    \State $\Bar{A} \gets \Delta \otimes \mathrm{parameter}^A$
    \State $\Bar{B} \gets \Delta \otimes B$
    \State $y_d \gets \mathrm{Selective Scan}(\Bar{A},\Bar{B},C)(s)$
\EndFor
\For{ \emph{d} in \{scan-directions\}}
    $y \gets y + y_d$
\EndFor
\State $y \gets \mathrm{LayerNorm}(y)$
\State $y \gets y * \mathrm{SiLU}(z)$
\State $y \gets \mathrm{Lin}(y)$
\end{algorithmic}
\end{algorithm}
\end{minipage}
\caption*{Comparison between the base VSSM block (on the left) and our ST-VSSM (on the right) that enables style injection.}
\end{figure*}
\setcounter{figure}{\value{tmp}}

\subsection{Style Transfer computation}
\label{sec:proof}
In this section we formalize the relation between style transfer computed via Transformers cross-attention and our Mamba-ST. In particular, we take inspiration from~\cite{ali2024hidden} and~\cite{dao2024transformers}, where a formalization of the relationship between Mamba and self-attention was first proposed. We start by noting that self-attention can be written as matrix transformation:
\begin{equation}
    Y = M X
    \label{eq:matrix_transformation}
\end{equation}
where $M = \mathrm{softmax}(QK^T)$ and $X = V$. Considering Mamba discretized equations:
\begin{equation}
     \begin{split}
     h_{k} &= \bar{A} h_{k-1} + \bar{B} x_k  \\        
     y_k &= C h_k + D x_k
     \end{split}
     \label{eq:mamba_supp}
\end{equation}
following \cite{ali2024hidden}, we can represent Eq.~\eqref{eq:mamba} as:
\begin{equation}
    y_t = C_t \sum_{j=1}^t\Big{(}{ \prod_{k=j+1}^t }\bar{A}_k\Big{)}\Bar{B}_j x_j
\end{equation}
To simplify the calculus, we remove $Dx_k$ since it can be seen as a skip connection multiplied by a scale factor $D$.
We can also define the matrices derivation as:

\begin{equation}
    \begin{split}
        C_i &= \mathrm{Lin_C}(x_i) \\
        \Delta_k &= \mathrm{softplus} \big{(}\mathrm{Lin_\Delta}(x_k)\big{)}\\
        \bar{A}_k &= \exp{\Big{(}\Delta_k A}\Big{)} \\
    \Bar{B}_j &= \Delta_k \mathrm{Lin_B}(x_j) 
    \end{split}
\end{equation}
Then, we  define $y_i=\sum_{j=1}^iy_{i,j}$, which maps the contribution of each input $x_j$ on the output $y_i$. In particular, $y_{i,j}$ is as follow:
\begin{equation}
\begin{split}
       y_{i,j} = & \underbrace{\mathrm{Lin_C}(x_i)}_{C}
            \rt{\Big{(}}\prod_{k=j+1}^i \underbrace{\exp{\bt{(}\overbrace{\mathrm{softplus} \gt{(}\mathrm{Lin_\Delta}(x_k)\gt{)}}^{\Delta_k} A\bt{)}}}_{\overline{A}_k}\rt{\Big{)}}\cdot \\
           & \underbrace{\overbrace{\mathrm{softplus} \gt{(}\mathrm{Lin_\Delta}(x_j)\gt{)}}^{\Delta_k}\mathrm{Lin_B}(x_j)}_{\overline{B}_j}x_j
\end{split}
\end{equation}
By simply applying exponential property we can rewrite it as:
\begin{equation}
\begin{split}
       y_{i,j} =& \underbrace{\mathrm{Lin_C}(x_i)}_{C}
            \rt{\Big{(}} \underbrace{\exp\bt{\Big{(}}\sum_{k=j+1}^i{\overbrace{\mathrm{softplus} \gt{(}\mathrm{Lin_\Delta}(x_k)\gt{)}}^{\Delta_k} A\bt{\Big{)}}}}_{\overline{A}_k}\rt{\Big{)}}\cdot \\
           & \underbrace{\overbrace{\mathrm{softplus} \gt{(}\mathrm{Lin_\Delta}(x_j)\gt{)}}^{\Delta_k}\mathrm{Lin_B}(x_j)}_{\overline{B}_j}x_j
\end{split}
\end{equation}
Furthermore, the two Softplus functions can be approximated summing only over positive values and using a ReLU:
\begin{equation}
    \begin{split}
        y_{i,j} \approx &  \underbrace{\mathrm{Lin_C}(x_i)}_{C}
            \Big{(}
                \underbrace{\exp{
                    \sum_{ 
                        \substack{k=j+1  \\ \mathrm{Lin_\Delta}(x_k)>0}
                   }^i }
                   \overbrace{\mathrm{Lin_\Delta}(x_k)}^{\Delta_k}
                   A }_{\overline{A}_k} \Big{)} \\
            & \underbrace{\overbrace{\mathrm{ReLU}(\mathrm{Lin_\Delta}(x_j))}^{\Delta_k}\mathrm{Lin_B}(x_j)}_{\overline{B}_j} x_j
    \end{split}
\end{equation}
Finally, considering the following values:
\begin{equation}
\begin{split}
    & Q_i = \mathrm{Lin_C}(x_i)) \\
    & K_j = \mathrm{ReLU}(\mathrm{Lin_\Delta}(x_j)\mathrm{Lin_B}(x_j))\\
    & H_{i,j} =  \exp{
                    \sum_{ 
                        \substack{k=j+1  \\ \mathrm{Lin_\Delta}(x_k)>0}
                   }^i }
                   \Big{(}
                   \mathrm{Lin_\Delta}(x_k)
                   \Big{)}A \Big{)}  \\
    & X = x_j
\end{split}
\end{equation}
We obtain that:
\begin{equation}
        Y = Q_iH_{i,j}K_{j} \cdot X
\end{equation}
which is a matrix transformation like \ref{eq:matrix_transformation}.
That demonstrates that we can simulate attention inside SSM, by assuming that $B,C$ and $\Delta$ play the role of Key, Query and Value from transformers attention, respectively.
Additionally, the matrix $A$ is the matrix that modulates the values of the previous patches, because is the one that is multiplied with the states $i-j$ inside the state space equations \ref{eq:mamba_supp}.

Finally, in order to compute style transfer in a similar way of StyTr2 \cite{deng2022stytr2}, but exploiting Mamba equations, we make the matrix $B,\Delta$, and consequentially $A$, dependent from the style $s$, and the matrix $C$ dependent from the content image $x$, employing a linear projection:
\begin{equation}
\begin{split}
 B = \mathrm{Lin_B}(s), \Delta = \mathrm{Lin_{\Delta}}(s), C = \mathrm{Lin_C}(x)
 \end{split}
 \label{eq:mamba_input_style_dependency_supp}  
\end{equation}
Additionally, we make the inner state, hence, the matrix $H_{i,j}$ style-dependent. By doing so, the output can be seen as a modulation of the inner state by the matrix $C$, which is content dependent.

\subsection{VSSM code comparison}
In this section we present the two versions of our ST-VSSM algorithm. On the left (alg.~\ref{alg:encoder}) is shown the Base VSSM algorithm used inside the two encoders in order to extract features from the images. As it can be seen, all the matrices depend only on the single input $x$ and are obtained with linear projections, as stated in \cite{liu2024vmambavisualstatespace}. On the right (alg.~\ref{alg:decoder}) the proposed ST-VSSM is shown, which is used to fuse the style and the content information in a single output. The first difference is that ST-VSSM takes as input both style and content information. It is worth noting that the depth-wise convolution shares the same weights for both the inputs. The core of ST-VSSM algorithm relies on how the matrices $A,B,C$ and $\Delta$ are computed: as it can be seen, the matrices $B$ and $\Delta$ (and consequentially the matrix $A$) are derived from the style source, while the matrix $C$ is derived from the content source. Moreover, the input of the selective scan is the style source and not the content. As stated, this is justified by the fact that we want to make the state dependent only from the style source, while modulating it with the content information in order to compute the output.

\begin{figure*}[!ht]
    \centering
    \includegraphics[width=\textwidth]{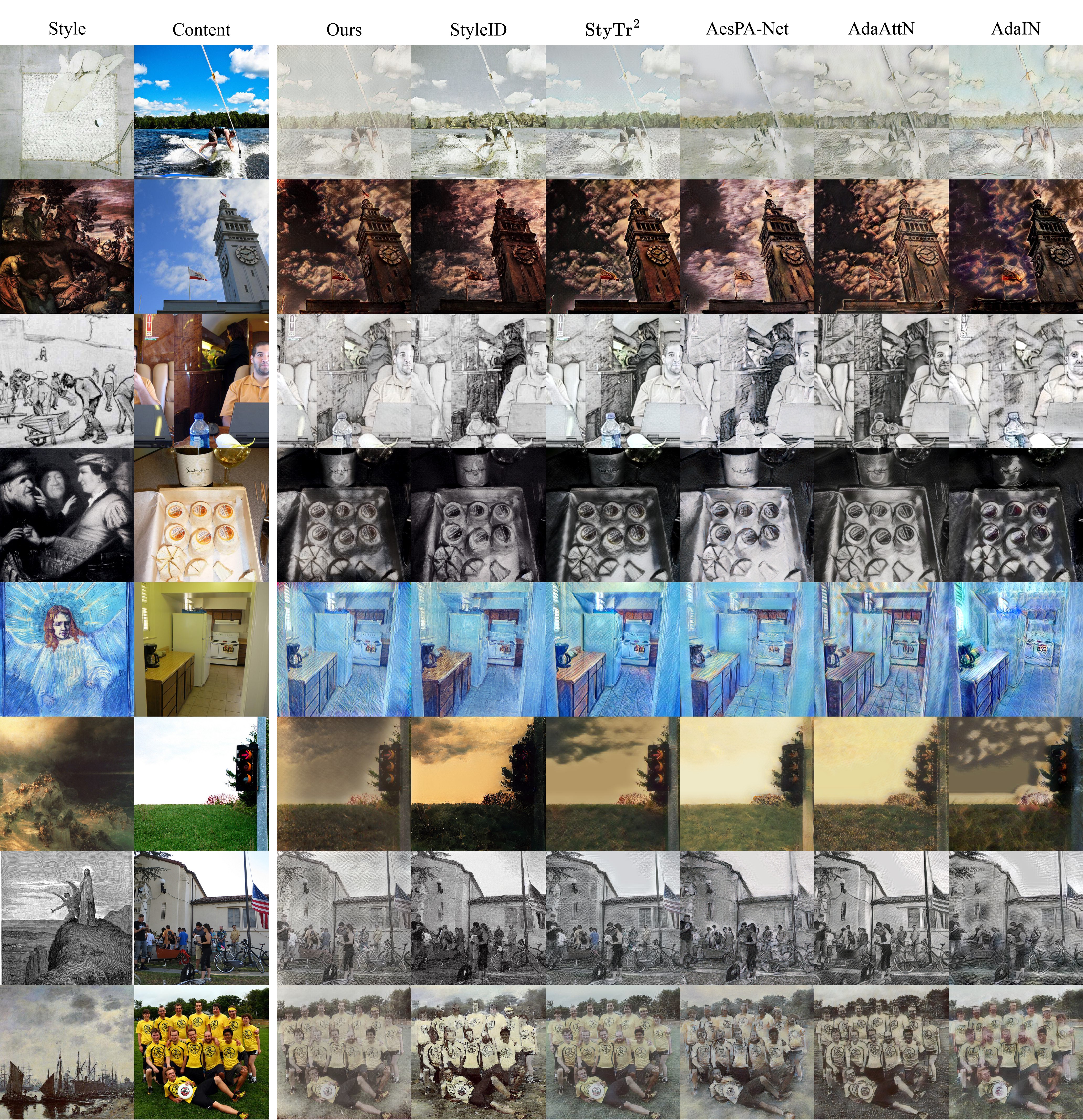}

   \caption{Additional comparisons with the current state-of-the-art models.}
    \label{fig:add_res}
\end{figure*}

\subsection{Additional Results}
In Fig. \ref{fig:add_res} we present additional results of our model compared with several state of the art architectures.

\end{document}